\documentclass[runningheads]{llncs}
\usepackage[T1]{fontenc}
\usepackage{graphicx}
\usepackage{ifthen}
\usepackage{xcolor}
\newboolean{showcomments}
\setboolean{showcomments}{true}

\ifthenelse{\boolean{showcomments}}
  {\newcommand{\nb}[3]{
  {\color{#2}\small\fbox{\bfseries\sffamily\scriptsize#1}}
  {\color{#2}\sffamily\small$\triangleright~$\textit{\small #3}$~\triangleleft$}
  }
  }
  {\newcommand{\nb}[3]{}
  }

\begin{document}
\title{Analysis Of The Anytime MAPF Solvers Based On The Combination Of Conflict-Based Search (CBS) and Focal Search (FS)\thanks{This is a preprint of the paper accepted to MICAI 2022}}
\author {
    Ilya Ivanashev \inst{1}, 
    Anton Andreychuk \inst{2}
    \and
    Konstantin Yakovlev \inst{3,1,2}
}

\institute{ 
    HSE University, Moscow, Russia \and
    AIRI, Moscow, Russia
    \and
    Federal Research Center for Computer Science and Control of Russian Academy of Sciences, Moscow, Russia\\
    \email{ivanashevi@yandex.ru, andreychuk@airi.net, yakovlev@isa.ru}
}

\authorrunning{I. Ivanashev et al.}

\titlerunning{Analysis Of The Anytime MAPF Solvers...}
\maketitle              

\begin{abstract}
Conflict-Based Search (CBS) is a widely used algorithm for solving multi-agent pathfinding (MAPF) problems optimally. The core idea of CBS is to run hierarchical search, when, on the high level the tree of solutions candidates is explored, and on the low-level an individual planning for a specific agent (subject to certain constraints) is carried out. To trade-off optimality for running time different variants of bounded sub-optimal CBS were designed, which alter both high- and low-level search routines of CBS. Moreover, anytime variant of CBS does exist that applies Focal Search (FS) to the high-level of CBS -- Anytime BCBS. However, no comprehensive analysis of how well this algorithm performs compared to the naive one, when we simply re-invoke CBS with the decreased sub-optimality bound, was present. This work aims at filling this gap. Moreover, we present and evaluate another anytime version of CBS that uses FS on both levels of CBS. Empirically, we show that its behavior is principally different from the one demonstrated by Anytime BCBS. Finally, we compare both algorithms head-to-head and show that using Focal Search on both levels of CBS can be beneficial in a wide range of setups.  

\end{abstract}
\section{Introduction}
Multi-agent pathfinding (MAPF) is a non-trivial problem that asks to find a set on collision-free paths for a set of mobile agents operating in the shared workspace. It naturally arises in robotics~\cite{VelosoBCR15}, video-games~\cite{Silver05}, automated warehouses~\cite{WurmanDM07}, aircraft-towing~\cite{MorrisPLMMKK16}, etc. One of the prominent MAPF solvers that is getting a lot of attention in the recent years is Conflict Based Search (CBS)~\cite{sharon2015conflict}. It is a two-level algorithm, which on the high level explores different solutions candidates and runs an individual planning for a specific agent on the low level.

CBS is a provably optimal MAPF solver that is highly modular in a sense that it allows modifications of its different algorithmic parts to either tweak its performance or adapt it to different MAPF problem statements. Indeed numerous enhanced modifications of CBS exists~\cite{CBSH2,lazy-cbs,andreychuk2019multi,li2019disjoint}. Still, the scalability of CBS is limited as it is tailored to find optimal MAPF solutions. On the other hand, practically-wise it is reasonable to trade-off optimality for lower runtime. This led the community to developing the bounded suboptimal versions of CBS~\cite{barer2014suboptimal,li2021eecbs}, i.e. such modifications of CBS that given a suboptimality factor $\varepsilon > 1$ return a solution whose cost does not exceed the cost of an optimal solution by a factor of $\varepsilon$ (the so-called $\varepsilon$-suboptimal solutions). While these algorithms are able to find a solution much faster than regular CBS, it might be difficult to choose the $\varepsilon$ value suitable for a particular MAPF problem. This problem is known to the community and one of the ways of solving it is designing \emph{anytime} versions of the algorithms, that gradually converge to an optimal solution via the series of bounded-suboptimal searches while keeping track of the found solutions (so the best available solution can be reported any time the algorithm is stopped).

We are aware of only one anytime bounded-suboptimal variant of CBS, that was recently proposed in~\cite{cohen2018anytime}. It combines CBS with the Focal Search~\cite{pearl1982studies} that is run on the high-level of CBS, while keeping the low-level search of CBS unchanged. In this work we will refer to this algorithm as Anytime BCBS, where BCBS stands for the bounded-suboptimal variant of CBS that keeps the low-level search unmodified as proposed in~\cite{barer2014suboptimal}. Anytime BCBS starts the search provided with the user-defined initial suboptimality factor (usually set rather high, e.g. $\varepsilon=10$) and iteratively decreases it while reusing the search efforts between the iterations. It was shown in the original paper~\cite{cohen2018anytime} that Anytime BCBS, indeed, converges to the very-close-to-optimal solutions under the strict time limits. However, it was not clear how Anytime BCBS compares to the naive sequential invocation of BCBS with the decreasing suboptimality factor without search re-use. This is the first research question we answer in this work, and the answer is that search re-use in Anytime BCBS is, indeed, beneficial.

Moreover, we suggest another variant of anytime MAPF
solver based on Conflict Based Search and Focal Search -- the one that utilize Focal Search on both levels of CBS. We call it Anytime ECBS, following the notation from~\cite{barer2014suboptimal}. Surprisingly, naive version of Anytime ECBS, i.e. the one that starts the search from scratch each time when the suboptimality factor is decreased, outperforms the version that reuses the search effort in many cases.

Finally, we compare the best version of Anytime BCBS, i.e. the one that relies on the search re-use, to the best version of Anytime ECBS, i.e. the one that runs the search from scratch on each iteration. Evidently, Anytime ECBS often  obtains the first solution faster and converges to the better quality solutions. Thus, one might infer that Anytime ECBS is a promising algorithm for the practical applications when the bounded-suboptimal solutions for challenging MAPF problems are sought under the strict time limits. 

\section{Problem Statement}
In classical MAPF~\cite{stern2019multi} one needs to find the set of collision-free paths for $n$ agents moving along the same graph $G=(V, E)$ from the specified start vertices to the goal ones. The time is discretized and at each time step an agent is allowed to perform one of the following actions: move to the neighboring vertex or wait in its current vertex. The duration of both actions are considered to be of the uniform (1 time step). The individual plan for each agent is defined as the sequence of actions that moves this agent from its start to goal. The cost of the plan is defined by the time step the agent reaches the goal. Plans for two agents are said to be conflict-free if the agents do not occupy any vertex and the same time step and do not traverse the same edge in the same time step. In other words they do not have any vertex or edge conflicts. 

MAPF solution is the set of individual plans for $n$ agents s.t. any pair of them is conflict-free. Cost of the MAPF solution is the sum-of-costs (SOC) of the individual plans comprising this solution. In this work we are interested in designing anytime bounded-suboptimal MAPF algorithm. I.e. the algorithm which is provided with \emph{i}) the MAPF instance, \emph{ii}) the time limit, \emph{iii}) the initial suboptimality bound $\varepsilon$, as an input, and is expected to initially find the $\varepsilon$-suboptimal solution very fast and use the remaining time for improving this solution, providing (when the time is up) the solution whose suboptimality bound is less than $\varepsilon$.

\begin{figure}[!t]
   \centering
   \includegraphics*[width=0.8\columnwidth]{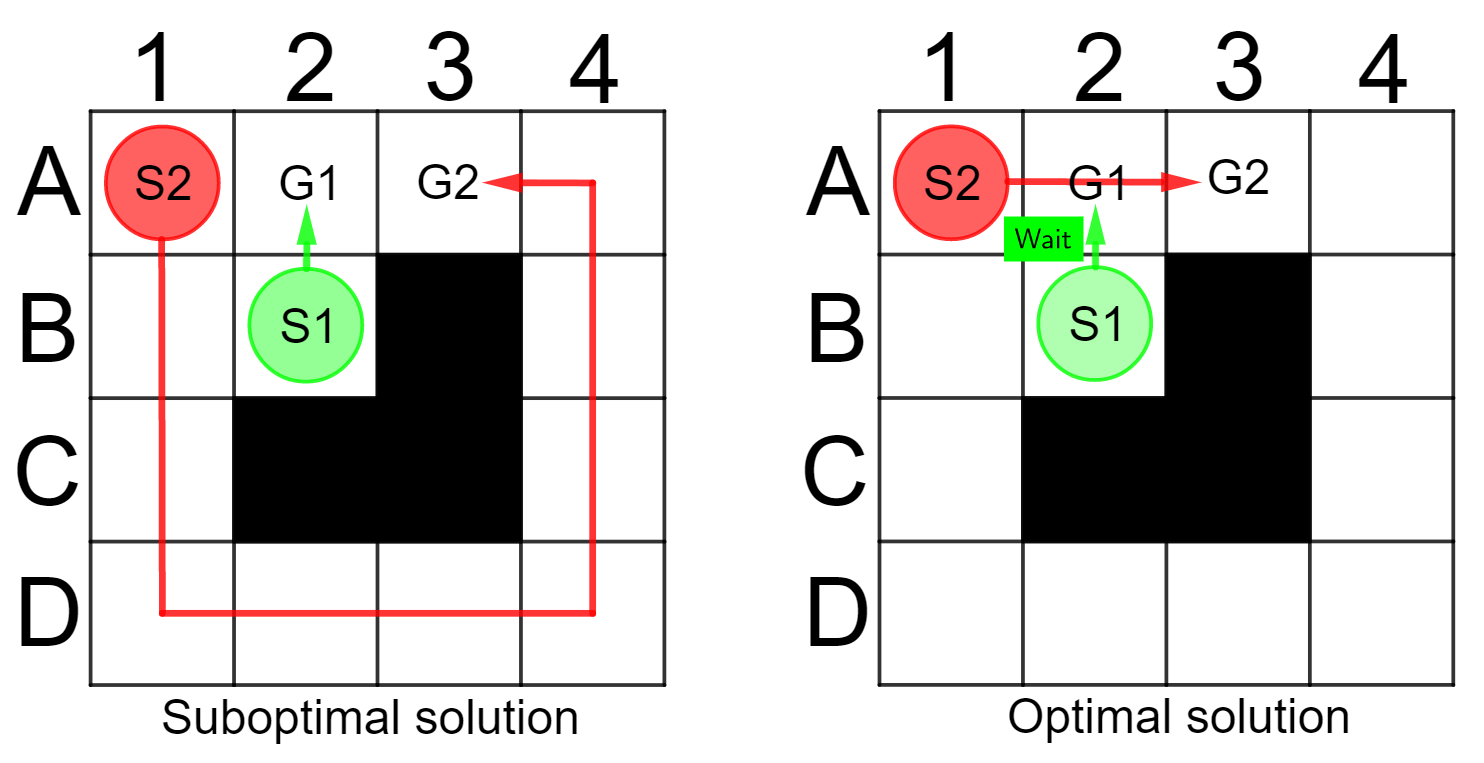}
   \caption{Examples of the suboptimal and optimal MAPF solutions.}
   \label{fig:Example.PNG}
\end{figure}

Figure~\ref{fig:Example.PNG} provides the example of MAPF instance with two possible solutions. 
Initially, if the $\varepsilon$ value is set high, an anytime algorithm can find a suboptimal solution, shown at the left part of the figure. In that solution the first agent immediately moves towards its goal node, blocking the way for the second agent, which then has to go around the obstacle. Then an algorithm can improve the obtained solution, finding an optimal one (shown in the right part of the figure), where the first agent waits during one time step, allowing the second agent to path by. 

\section{Background And Related Work}
\paragraph{Conflict Based Search}
Conflict-Based search (CBS)~\cite{sharon2015conflict} is an optimal MAPF solver which operates on two levels. The high level of the algorithm builds the Constraints Tree (CT). Each node in this tree (CT-node) is characterized by the set of constraints (that tell which vertices/edges particular agents should not occupy in which time steps), the set of individual plans (consistent with the constraints), and the cost (the sum of costs of individual plans). The root node contains an empty set of constrains and the set of individual plans that are likely to contain conflicts. On each step the algorithm chooses a CT-node with the least cost and checks it for conflicts. If the chosen node has no conflicts, CBS reports finding the optimal solution. Otherwise, the algorithm picks a conflict and creates two successor CT-nodes whose constraints sets are extended by one constraint in the following way. If the picked conflict was between the agents $i$ and $j$ which visited the vertex (edge) $v$ ($e$) at time $t$, then the first successor is provided with the constraint, prohibiting $i$ from being at $v$ ($e$) at time $t$, and the similar constraint but for the agent $j$ is added to the second successor. The individual plans of the constrained agents are then rebuilt with a low-level search algorithm (to be discussed later) and the CT-nodes are added to the CT-tree. CBS now proceeds to picking the next best CT-node. 

Numerous extensions of CBS exist nowadays that boost its performance while not violating the optimality guarantees. Among the most widespread techniques are: prioritizing conflicts~\cite{sharon2015conflict}, by-passing conflicts~\cite{boyrasky2015don}, adding heuristics to high-level~\cite{felner2018adding}, disjoint splitting~\cite{li2020new} and many others. As in this work we are not targeting optimal solutions (but rather bounded sub-optimal), these improvements are not within the scope of the paper.

\paragraph{Low-Level Planning for CBS}
The basic version of CBS uses A*~\cite{hart1968formal} as the low level planner. The search is performed in the space where each state is characterized by a pair $(v, t)$ where $v$ is a graph vertex and $t$ is a time step. When expanding a search node $(v, t)$ the successors corresponding to moving to the adjacent vertices are added to the search tree (if transitions to them are not prohibited by the CBS constraints). Moreover, the successor corresponding to the wait action, i.e. $(v, t+1)$, is also added as well (if not prohibited by CBS constraints).

A technique that is frequently used for the A* search operating at the low-level of CBS is breaking ties in accordance with the \emph{Conflict avoidance table} (CAT). This structure stores the information about the number of agents which visit every vertex and move through every edge at any time step. Thus, when similarly perspective nodes are encountered by A* (i.e. the ones that have the same $f$-value) the algorithm picks the one that have the lowest number of conflicts in CAT.

Another popular low-level planner for CBS is Safe Interval Path Planning (SIPP)~\cite{phillips2011sipp}. It operates with the time intervals as opposed to distinct time steps and is notably faster than A*. Recently, even more advanced variants of low-level planning for CBS were proposed~\cite{hu2022multi}. However, in our work we mainly focus on the basic variant of the low-level search for CBS, i.e. A* with CAT, as it is more straightforward (and widespread) to integrate it with the bounded-suboptimal variants of CBS.

\paragraph{Focal Search}
Focal search~\cite{pearl1982studies} is a variant of the A* algorithm that is tailored to produce $\varepsilon$-suboptimal solutions. Besides conventional OPEN and CLOSED lists used in A*, Focal Search uses an additional FOCAL list that contains a subset of nodes from OPEN with $f$-values that differ from the minimal $f$-value no more than by a factor of $\varepsilon$. All the nodes in FOCAL list are ordered in accordance with a heuristic $h_{focal}$, which is not needed to be monotone or consistent. Focal Search can be utilized within the CBS framework to obtain a bounded-suboptimal MAPF solver as described below.

\paragraph{Bounded-Suboptimal Variants of CBS}
Different variants of bounded-suboptimal CBS exist. The most relevant to this work are the ones introduced in~\cite{barer2014suboptimal}, i.e. Bounded CBS (BCBS) and Enhanced CBS (ECBS).

The first variant uses the Focal Search both on high and low levels of CBS with two separate bounds ($\varepsilon_H$, $\varepsilon_L$). On the low level BCBS uses $h_{focal}$  which is defined as the number of conflicts accumulated by the path from the start node to the current one, which is obtained from CAT. For the the high level $h_{focal}$ is defined as the total number of conflicts in CT-nodes (other variants also possible). Contrary to BCBS, Enhanced CBS allows to specify a single suboptimality bound $\varepsilon$ and distribute it between high and low levels automatically. It was shown empirically that ECBS($\varepsilon$) outperforms BCBS with different combinations of $\varepsilon_H$ and $\varepsilon_L$, s.t. $\varepsilon_H \cdot \varepsilon_L=\varepsilon$, and in particular BCBS($\varepsilon$, 1).

More advanced variants of bounded-suboptimal CBS appeared recently~\cite{li2021eecbs,chan2021ecbs}. In this work, however, we focus on BCBS and ECBS and use them as building blocks for Anytime CBS. Implementing Anytime CBS using the advanced variants of bounded-suboptimal CBS is an appealing direction for future research. 

\paragraph{Anytime Variants of CBS}
The easiest way of creating an anytime version for a $\varepsilon$-suboptimal algorithm consists in the following naive approach. During the first iteration a regular version of the algorithm is being run, with an initial value of $\varepsilon$. After the solution is encountered, $\varepsilon$ is decreased in such a way that this solution does not meet it. Thus the new solution that meets the bound has to be found, and the algorithm is restarted from scratch with this decreased value of $\varepsilon$. The algorithm continues to perform such iterations with a decreasing sequence of $\varepsilon$ values, until it has some time left for execution or until it is able to find an optimal solution.

More advanced anytime variant of CBS, which is able to re-use the results of previous iterations, was presented in~\cite{cohen2018anytime}. It builds on the anytime modification of Focal Search (AFS) and applies this technique to the high level of the BCBS($\varepsilon$, 1) algorithm, while keeping its low-level (A* + CAT) unchanged. We call this algorithm Anytime BCBS. As in the naive approach, in Anytime BCBS algorithm BCBS($\varepsilon$, 1) is being run iteratively with the decreasing suboptimality bound $\varepsilon$, where the new value for $\varepsilon$ is always chosen in such a way that it is not being met by the old solution. The difference, however is that the algorithm does not start growing the new CT-tree from scratch but rather continues to grow the CT-tree constructed in the previous iteration. In order to achieve that, all nodes in the high-level FOCAL list that no longer satisfy the $\varepsilon$-suboptimality condition are removed from it, after which the search process resumed. Intuitively, this saves computational effort, however the original paper on Anytime BCBS lacks comparison with the naive approach, described above. In this work we fill this gap and empirically show that, indeed, re-using the CT-tree for Anytime BCBS is beneficial.

It should be noted here that numerous other (non CBS-based) anytime MAPF planners exist nowadays. The most prominent are: MAPF-LNS~\cite{li2021anytime} and MAPF-LNS2~\cite{li2022mapf} etc. These planners, however, do not provide any theoretical guarantees on the cost of the returned solution, while we in this work are interested in getting bounded-suboptimal solutions.

\section{Anytime ECBS}
As with the BCBS algorithm, one can construct the anytime version for ECBS using the naive approach. However, we have also developed a version of the algorithm, referred to as Anytime ECBS, which is able to reuse the results of the previous iterations. It works as follows. The first iteration of the algorithm invokes ECBS and obtains a $\varepsilon$-suboptimal solution. The only difference is that OPEN$_N$, FOCAL$_N$ and CLOSE$_N$ lists of the low level planner are saved for every CT-node $N$. Further these lists will be used to efficiently rebuild the individual agents' trajectories between the iterations with respect to the new value of $\varepsilon$. This have to be done, because, in contrast with BCBS($\varepsilon$, 1), these trajectories are also suboptimal and depend on the current value of suboptimalty factor. 

After decreasing the value of $\varepsilon$ to the lower value $\varepsilon'$, a new search iteration is performed. For each CT node $N$ including already expanded CT nodes, we update the trajectories with respect to the new value of $\varepsilon'$ (if necessary) using the saved OPEN$_N$, FOCAL$_N$ and CLOSE$_N$ lists. To do that, for every high-level node $N$ a new AFS iteration is performed: all low-level nodes which don't meet the new suboptimality factor, are removed from FOCAL$_N$, and the low level FS is resumed and continued until an $\varepsilon'$-optimal solution is found. This procedure is applied to CT nodes downward from the root to the leafs of CT tree. Thus, when any CT node is updated, agents' paths in its predecessors are already rebuilt and added into CAT, so the new path can be constructed with minimal amount of conflicts with other trajectories. However, it should be mentioned, that the values of $h_{focal}$ heuristic aren't recalculated for the nodes, which were already added into the OPEN$_N$, FOCAL$_N$ and CLOSE$_N$ lists at the end of previous iteration, as it would require to traverse the whole low level search tree again. Because of that some conflicts may be created, which could be otherwise avoided, if the trajectory was constructed from scratch. That can have negative effect on the performance of Anytime ECBS and may be one of the potential reasons, why it would expand more high level nodes, than the naive algorithm for some problems. However, it is hard to tell how much influence this issue had in the experiments, conducted in our study, as well as to fully explain the differences in behavior between Anytime ECBS and its naive version. The reason for this is that the sequences of CT nodes expanded by the algorithms can diverge at the very beginning of the search and become completely different, after which it is hard to determine, why a particular algorithm was able to find a better solution, or finish the execution earlier.

The constraints sets in all the nodes remain the same and have to be satisfied in the updated trajectories. It is possible that some constraints in constraint set will become irrelevant, i.e. applying them will not help to prevent any conflicts, because trajectories of the agents, which were involved in the conflict previously, have already been changed. 

We consider two possible strategies of dealing with such situations. Firstly, one can remove the whole subtree of this node from the CT, and reinsert it into the OPEN list in order to find a new conflict in the solution, which corresponds to it. Alternatively, all nodes can be left in the CT with the same constraints. The initial hypothesis, prompting us to apply the second strategy, was that even irrelevant constraints can still be useful, because they are often applied for the positions with high probability of conflict appearance and therefore they may help to prevent some future conflicts (this hypothesis however wasn't actually confirmed by the experiments, and it was seen that keeping irrelevant constraints in the constraints set, usually only worsens algorithms performance). We denote the parameter, defining what strategy to use, as CIC (cut irrelevant conflicts): if CIC = True the first strategy is used and if CIC = False, the second.

After all CT nodes are updated, every node contains a $\varepsilon'-$suboptimal solution satisfying all of the constraints in the node's constraint set. Then the nodes with the solution costs that do not meet the new threshold are removed from the high-level FOCAL list. After that the search is continued until a new solution without conflicts is found. 

\paragraph{Example}
\begin{figure}[!t]
   \centering
   \includegraphics*[width=\columnwidth]{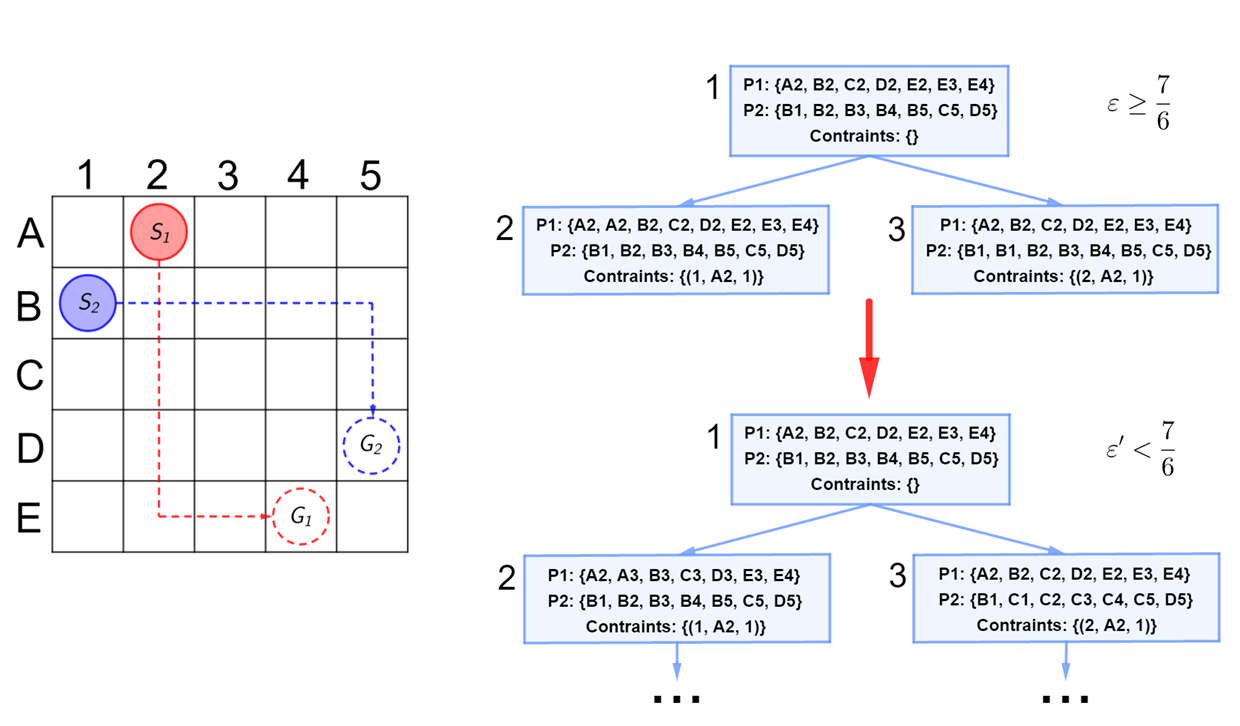}
   \caption{Running example of Anytime ECBS on a simple MAPF problem.}
   \label{fig:example}
\end{figure}

This section shows how the CT is updated between the stages of anytime ECBS algorithm. In the MAPF problem, shown on the Figure \ref{fig:example} there are two agents with starting nodes $S_1=A2$, $S_2=B1$ and goal nodes $G_1=E4$, $G_2=D5$. Their initial individual trajectories, shown with dotted lines, have cost 6 and have a conflict in the node $B2$ at the moment 1. If the intial value of $\varepsilon$ is not less than $\frac{7}{6}$, the first iteration of Anytime ECBS will be able to find a valid solution in both successors of the root node, by adding one waiting action into the agents trajectories. The whole CT, which will be constructed by the algorithm in this case, is shown at the right top of the figure (constraints are presented as $(a, n, t)$ triples, meaning that agent $a$ is prohibited from being in node $n$ at time $t$). In contrast, Anytime BCBS algorithm would have to create more high-level nodes in the first iteration, because updated agents trajectories would still be optimal and a new conflict would appear (e.g. the first agent would satisfy the constraint by firstly moving to A3 and then moving down, but it would create a conflict in the node B3). 

After the end of the first iteration, $\varepsilon$ will be replaced with new suboptimality factor $\varepsilon'<\frac{7}{6}$. Then the trajectories of agents 1 and 2 in the CT nodes 1 and 2 will be rebuild in order to satisfy the new suboptimality factor, which means, that they will have to become optimal trajectories of length 6 (see the right bottom of the figure). Then, similarly to the first iteration of Anytime BCBS, algorithm would have to add new constraints to rediscover a valid solution again.

\section{Empirical evaluation}

\begin{figure}[!t]
    \centering
    \includegraphics[width=\columnwidth]{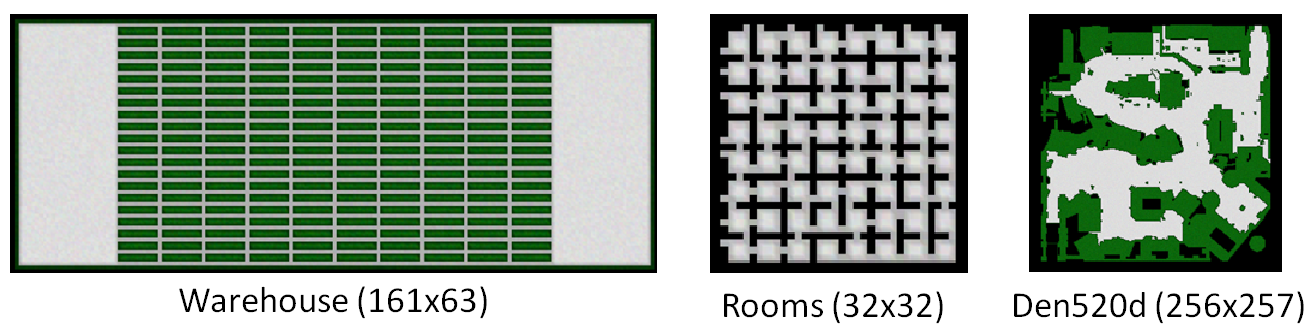}
    \caption{Maps used for the evaluation.}
    \label{fig:maps.png}
\end{figure}

The algorithms were implemented in C++ (the code can be found at~\footnote{\url{https://github.com/PathPlanning/Push-and-Rotate-{}-CBS-{}-PrioritizedPlanning}}) and evaluated on 4 different maps taken from the well-known in the MAPF community MovingAI benchmark~\cite{stern2019multi}: \texttt{empty-16-16}, \texttt{room-32-32-4}, \texttt{warehouse-10-20-10-2-2} and \texttt{den520d} (for the sake of simplicity later in the article we refer to them as \texttt{Empty}, \texttt{Warehouse}, \texttt{Rooms} and \texttt{Den520d}). These maps are widely used to evaluate MAPF algorithms as they represent different types of the environments as shown on \figurename~\ref{fig:maps.png} (empty map is not shown, obviously).

For every map the benchmark provides $25$ distinct scenarios. Each scenario is a list of (non-overlapping) start-goal locations. To obtain a MAPF instance for $k$ agents first $k$ start-goal pairs form the scenario are used. While evaluating we incrementally increase the number of agents to obtain the instances of the variable difficulty. 

The experiments were run on Intel Core i5-1135G7 CPU, 2.40GHz, 4 cores running Windows 10 Home OS. We imposed a time limit of $90$ seconds for solving each instance, i.e. if an algorithm was not able to find a solution it was interrupted and this run was counted as failure.

\begin{figure}[!t]
    \centering
    \includegraphics[width=1.0\columnwidth]{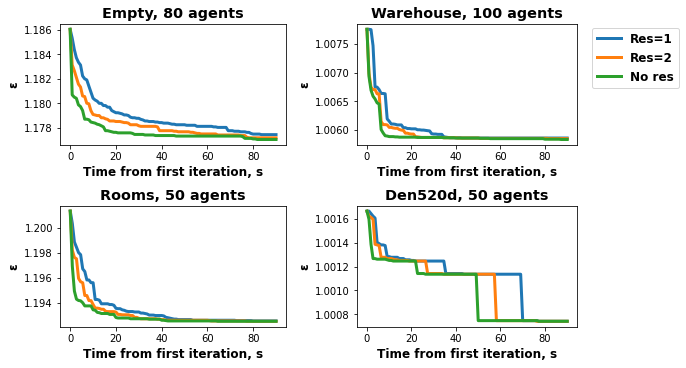}
    \caption{Changes of $\varepsilon$ over time for different versions of Anytime BCBS algorithm.}
    \label{fig:BCBS_eps_len.png}
\end{figure}

\paragraph{Evaluating Anytime BCBS.} In the first series of experiments we compared three versions of the Anytime BCBS algorithm (dubbed as ABCBS):
\begin{itemize}
    \item ABCBS, res=1 -- the algorithm that always starts a new search from scratch, i.e. never re-uses the previously built CT-tree;
    \item ABCBS, res=2 -- the algorithm that starts the search from scratch on odd iterations, and on even iterations re-uses the previously build CT-tree;
    \item ABCBS, No res -- the algorithm that always continues to use the previously built CT-tree.
\end{itemize}

The initial value of $\varepsilon$ was set to $10$ to increase the chance of finding the first (possibly highly suboptimal) solution quickly. At each following iteration $\varepsilon$ was decreased in the way described in Section 3. \figurename~\ref{fig:BCBS_eps_len.png} shows how $\varepsilon$ was changing for certain setups (map + number of agents). 

Each point corresponds to the suboptimality factor $\varepsilon$, for which the solution was already found, averaged over all of the scenarios for which the algorithm was able to find at least one solution within the given time limit. It also should be noted that for different scenarios the completion time of the first iteration could be different, so the $x$ axis shows the relative time starting from that moment.

\begin{figure}[!t]
    \centering
    \includegraphics[width=0.7\columnwidth]{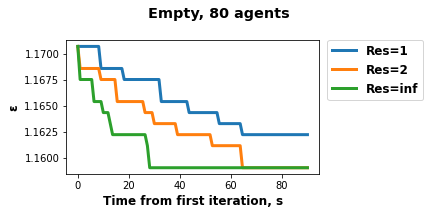}
    \caption{Comparison of different versions of Anytime BCBS on particular scenario.}
    \label{fig:BCBS_empty_80_len.png}
\end{figure}

As one can see, the version of the algorithm, which is never restarted, shows the best results out of all of the versions. This can be explained by the fact, that the sequences of $\varepsilon$ values are approximately the same for all versions of the algorithm, as can be seen, for example, on  Figure~\ref{fig:BCBS_empty_80_len.png}, that shows how the value of $\varepsilon$ changes over time on one particular scenario. As result of that, the sizes of CT trees at the end of every iteration are also close for different versions, and version without restarts is able to finish every new iteration faster, compared to the naive algorithm, since it doesn't need to rebuild the first part of constraint tree.

Similar comparison was also performed for Anytime ECBS algorithm. However, it turned out, that setting initial value of $\varepsilon$ too high can actually slow the ECBS algorithm down, because it would have to expand more nodes on the low level. For example, the following situation can be considered. Let's say that in some intermediate solution in ECBS algorithm, agent $i$ stops in its goal node $n$ at the moment $t$. Also let's say that agent $j$ has to go through the node $n$ at the moment $t' > t$. Then there will always be a conflict with an agent $i$ in its trajectory in the node $n$. If the value of $\varepsilon$ is set too high, the algorithm will spend a lot of time, trying to avoid this conflict, and expanding a lot of low level nodes. That makes the low level search very time consuming, which negates the possible time saving from expanding less nodes on the high level.

Considering the observations, described above, it was decided to specifically select a preferable initial value of $\varepsilon$ for Anytime ECBS for every map.

After running ECBS using different values of $\varepsilon$ on tasks with different numbers of agents for all four maps, we have found that the best results were obtained when using the $\varepsilon$ value 2 for the maps \texttt{Empty} and  \texttt{Rooms}. For the maps \texttt{Warehouse} and \texttt{Den520d} results for all considered values of $\varepsilon$ were very close, so the value 10 was used, as in Anytime BCBS.

\begin{figure}[!t]
    \centering
    \includegraphics[width=1.0\columnwidth]{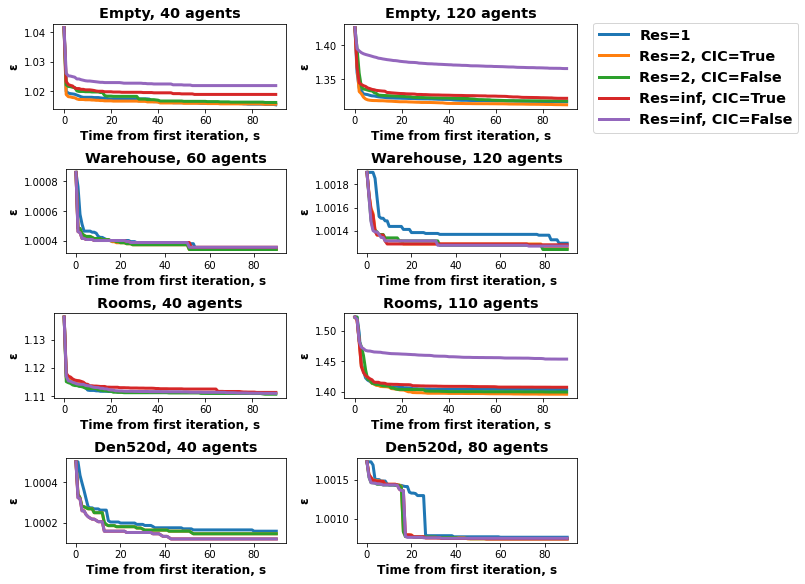}
    \caption{Changes of $\varepsilon$ over time for different versions of Anytime ECBS algorithm.}
    \label{fig:ECBS_eps_len_2.png}
\end{figure}

Then the comparison of different Anytime ECBS versions was performed using the selected initial values of $\varepsilon$. The results are provided in Figure~\ref{fig:ECBS_eps_len_2.png}. For every map the results for two different agent numbers are shown. It's also should be mentioned that for the \texttt{Rooms} map in 2 scenarios there were less than 110 agents provided in the data set, thus these scenarios weren't considered for the corresponding plot. 

In addition to the Res parameter, previously described parameter CIC was considered. That gives us 5 possible configurations of parameters (for the naive version of the algorithm only CIC=True is possible, because all CT is removed after every iteration). The results depicted in Figure~\ref{fig:ECBS_eps_len_2.png} indicate that in many cases the naive version of Anytime ECBS algorithm allows to obtain solutions with approximately the same or even better quality as the advanced versions. This can be explained by the fact, that the decreasing sequences of $\varepsilon$ values in that algorithm can vary significantly between different versions, and the advanced version may have to expand more high level nodes than the naive one. Alternatively, for some scenarios the number of nodes, expanded by the advanced version, could be approximately the same, or even less, but the value of $\varepsilon$ would decrease slower.

As was mentioned above, the naive version of Anytime ECBS can potentially get an advantage over advanced version, because in the latter algorithm values of $h_{focal}$ heuristic aren't updated between iterations. Moreover, the performance of Anytime ECBS can be negatively affected by keeping the irrelevant constraints in the versions of the algorithm with CIC=False. In particular, the version of the algorithm that never restarts (and therefore can have the least relevant values of $h_{focal}$ heuristic) and keeps the irrelevant constraints shows considerably worse results for the maps \texttt{Rooms} and \texttt{Empty} in the tasks with high number of agents. Although, as it was previously mentioned, it is not clear, how much influence each of these points had, and the difference between the versions on the instances with lower number of agents was even less noticeable.

\begin{figure}[!t]
    \centering
    \includegraphics[width=0.8\columnwidth]{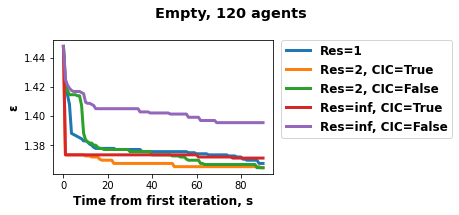}
    \caption{Comparison of different versions of Anytime EBCBS on particular scenario.}
    \label{fig:ECBS_empty_120_len.png}
\end{figure}

Chart for a particular scenario on \texttt{Empty} map, shown on Figure~\ref{fig:ECBS_empty_120_len.png}, indicates that in contrast to Anytime BCBS there is almost no correlation between the decreasing sequences of $\varepsilon$ values for different versions of Anytime ECBS algorithm.

However, for two other maps -- \texttt{Warehouse} and \texttt{Den520d}, advanced versions of the algorithm show a noticeable improvement over the naive one for some of the tasks with the higher number of agents, although there was no clear leader between them. Nevertheless, the fact that the naive version doesn't require additional memory consumption and is easier to implement, means that it probably can be preferred over the other versions.

\begin{figure}[!t]
    \centering
    \includegraphics[width=1.0\columnwidth]{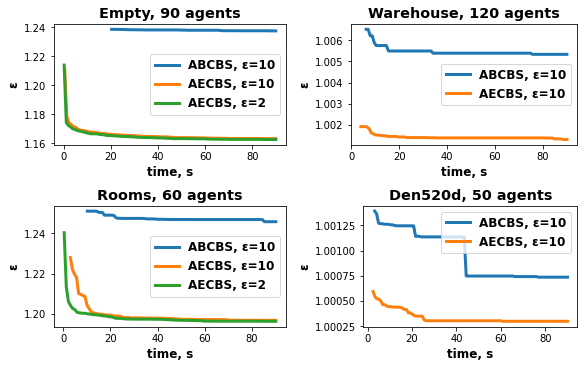}
    \caption{Comparison between anytime BCBS and Anytime ECBS.}
    \label{fig:BCBS_vs_ECBS_3_len.png}
\end{figure}

Finally, the version of Anytime BCBS algorithm without restarts was compared to the naive version of Anytime ECBS algorithm. Additionally, for the maps \texttt{Empty} and \texttt{Rooms}, where a specifically chosen value of $\varepsilon$ was used, a version with initial value of $\varepsilon$ equal 10 was added to the comparison. The results of this experiment are shown in Figure~\ref{fig:BCBS_vs_ECBS_3_len.png}. As one can see, Anytime ECBS often finds better solutions than Anytime BCBS. This can be a result of higher flexibility of ECBS algorithm: it is able to build slightly worse trajectory for one agent, in order to avoid the creation of new conflicts, which would then allow to make better trajectories for the other agents (although, as in the previous cases it might be not the full explanation and might be not relevant for some of the problems). Considering that and the fact that ECBS algorithm is usually faster than BCBS($\varepsilon$, 1), i.e. the first iteration in Anytime ECBS can be finished faster, naive version of Anytime ECBS can be preferred to the advanced version of Anytime BCBS. The one can also note that on the \texttt{Rooms} map the version of Anytime ECBS with $\varepsilon$ equal 10 was slower, and found the initial solution later, while for the \texttt{Empty} map both versions of the algorithm were able to find an initial solution very fast, so there wasn't much difference between them. 

It should be also mentioned, that for some maps and algorithms there were scenarios where different algorithms were not capable to find even initial solution within the timelimit. In addition to that, for some scenarios algorithms were only able to finish one iteration, and therefore such scenarios didn't give any additional information about their anytime properties. Particularly, large number of such scenarios was presented for the Anytime BCBS algorithm on the \texttt{Warehouse} and \texttt{Den520d} maps. For example, there were only 8 scenarios for the \texttt{Warehouse} map with 100 agents and only 11 for the \texttt{Den520d} map with 50 agents, where Anytime BCBS was able to finish more than one iteration. 

\section{Conclusion and Future Work}

In this work we have presented a novel bounded-suboptimal anytime MAPF solver, based on the prominent ECBS algorithm - Anytime ECBS. We empirically compared it with previously existing bounded-suboptimal anytime MAPF solver -- Anytime BCBS. Both algorithms were also compared with their naive versions, in which the search is restarted from scratch at the beginning of every iteration. It was shown, that while Anytime BCBS always outperforms its naive version, for Anytime ECBS the naive algorithm can often be preferred. However, even the naive version of Anytime ECBS had some advantages over the Anytime BCBS algorithm, as the former was usually able to finish the first iteration faster, and the solutions obtained by it typically had lower costs compared to the ones found by Anytime BCBS.

Avenues for future work may include developing novel techniques aimed at decreasing the number of low level nodes which have to be stored in Anytime ECBS, or investigation of how the stored information can be used to efficiently rebuild agents paths after addition of the new constraints not only during the process of repairing the constraint tree between the iterations of anytime algorithm but also during its building (analogously to incremental search techniques).

\bibliographystyle{splncs04}
\bibliography{references}

\end{document}